\pdfoutput=1

\documentclass{article}
\usepackage{ijcai20}

\usepackage{times}

\usepackage{soul,xcolor}
\usepackage{url}
\makeatletter
\g@addto@macro{\UrlBreaks}{\UrlOrds}
\makeatother
\usepackage[hidelinks]{hyperref}
\usepackage[utf8]{inputenc}
\usepackage[small]{caption}
\usepackage{graphicx}
\usepackage{amsmath}
\usepackage{booktabs}
\usepackage{amsfonts}
\title{A Review of Winograd Schema Challenge Datasets and Approaches}

\author{
Vid Kocijan$^1$\footnote{Contact Author}\and
Thomas Lukasiewicz$^{1,2}$
\and Ernest Davis$^3$\and
Gary Marcus$^4$\And
Leora Morgenstern$^5$\\
\affiliations
$^1$University of Oxford\\
$^2$Alan Turing Institute, London\\
$^3$New York University\\
$^4$Robust AI \\
$^5$Systems \& Technology Research / PARC
\emails
firstname.lastname@cs.ox.ac.uk,
davise@cs.nyu.edu,
gary.marcus@nyu.edu,
leora.morgenstern@gmail.com
}

\begin{document}

\maketitle

\begin{abstract}
The Winograd Schema Challenge 
is both a commonsense reasoning and natural language understanding challenge, introduced as an alternative to the Turing test.
A Winograd schema is a pair of sentences differing in one or two words with a highly ambiguous pronoun, resolved differently in the two sentences, that appears to require commonsense knowledge to be resolved correctly.
The examples were designed to be easily solvable by humans but difficult for machines, in principle requiring a deep understanding of the content of the text and the situation it describes.
This paper reviews existing Winograd Schema Challenge benchmark datasets and approaches that have been published since its introduction.
\end{abstract}

\section{Introduction}

The Winograd Schema Challenge
was introduced by Hector Levesque~\cite{WinogradSchema}  both as an alternative to the Turing Test  \cite{TT} and as a test of a system’s ability to do commonsense reasoning.

An example of a Winograd schema is the pair of sentences introduced by Terry Winograd~\shortcite{WinogradUnderstandingNL}:\smallskip\\
 \textit{The city councilmen refused the demonstrators a permit because \textbf{they} [feared/advocated] violence.}\\
\textbf{Question: } Who [feared/advocated] violence?\\
\textbf{Answer: } the city councilmen / the demonstrators

\smallskip 
The word \textit{they} refers to the city councilmen or the demonstrators, depending on whether the word \textit{feared} or \textit{advocated} is used in the sentence. To correctly identify the referent, a human would probably need to know a good deal about demonstrators, permits, city councilmen, and demonstrations.

Levesque’s insight was that one can construct many other pairs of sentences, which appear to rely on commonsense reasoning, and for which sentence structure does not help disambiguate the sentence.  He claimed that a system that could achieve human performance in solving such sentences would have the commonsense knowledge that humans use when doing such disambiguation. Such pairs of sentences would have to be constructed to have certain properties, including being identical except for one or two ``special’’ words and not be solvable by  \textit{selectional restriction}.  An important constraint was that the Winograd schemas be ``Google-proof'' or non-associative~\cite{WSCAnalysis}, meaning that one could not use statistical associations to disambiguate the pronouns.
As we discuss below, this is the least achievable constraint, as indicated by the success of statistical language models described in the survey.

The Winograd Schema Challenge was appealing, because the task of pronoun disambiguation is easy and automatic for humans, the evaluation metrics were clear, and the trick of using twin sentences seemed to eliminate using structural techniques to get to the right answer in ways that avoided using commonsense reasoning.  In the years following its publication, the challenge became a focal point of research for both the commonsense reasoning and  natural language processing 
communities. 

A great deal of progress has been made in the last year. In this paper, we review the nature of the test itself, its different benchmark datasets, and the different techniques that have been used to address~them.

\section{Winograd Schema Challenge Datasets}
Several Winograd Schema Challenge datasets have been introduced;  for the most part, they can be split into two categories: performance-measuring and diagnostic datasets.


\subsection{Original Collection of Winograd Schemas}

The first collection of $100$ Winograd schemas were published together with the introduction of the Winograd Schema Challenge~\cite{WinogradSchema}\footnote{\url{https://cs.nyu.edu/faculty/davise/papers/WinogradSchemas/WS.html}}.
Examples are constructed manually by AI experts, with the exact source for each example available.
At the time of writing, there are $285$ examples available; however, the last $12$ examples were only added recently.
To ensure consistency with earlier models, several authors often prefer to report the performance on the first $273$ examples only.
These datasets are usually referred to as \textsc{Wsc285} and \textsc{Wsc273}, respectively.

\paragraph{Subclasses of the original collection}

Trichelair \emph{et al.}~\shortcite{WSCAnalysis} have observed that $37$ sentences in the \textsc{Wsc273} dataset ($13.6\%$) are conceptually easier than the rest. 
The correct candidate is commonly associated with the rest of the sentence, while the incorrect candidate is not.
An example of such a sentence is 

\textit{In the storm, the tree fell down and crashed through the roof of my house. Now, I have to get \textbf{it} [repaired/removed].}

\textit{The roof} is commonly associated with \textit{being repaired}, while \textit{the tree} is not.
They call these examples \textit{associative} and name the rest \textit{non-associative}.
Moreover, they find that models often perform much better on the associative subsets.

Additionally, $131$ sentences ($48\%$ of \textsc{Wsc273}) were found to form meaningful examples if the candidates in the sentence are switched.
An example of such sentence is 

\textit{Bob collapsed on the sidewalk. Soon he saw Carl coming to help. \textbf{He} was very [ill/concerned].}

In this sentence, \textit{Bob} and \textit{Carl} can be switched to obtain an equivalent example with the opposite answers.
Such sentences were named \textit{switchable}.
Trichelair \emph{et al.}~\shortcite{WSCAnalysis} encourage future researchers to additionally report the consistency on the \textit{switchable} dataset, when the candidates are switched, and when they are not.

The list of associative and switchable examples together with their switched counterparts have been made public\footnote{\url{https://github.com/ptrichel/How-Reasonable-are-Common-Sense-Reasoning-Tasks}}.

\paragraph{Winograd Schema Challenge in other languages.}

While the inspiration and original design of the challenge was in English, translations into other languages exist.
Amsili and Seminck~\shortcite{WSCfrench} translated the collection of $144$ Winograd schemas into French, and $285$ original Winograd schemas were translated into Portugese by Melo \emph{et al.}~\shortcite{WSCportugese}.
Additionally, translations to Japanese\footnote{\url{http://arakilab.media.eng.hokudai.ac.jp/~kabura/collection_katakana.html}} and Chinese\footnote{\url{https://cs.nyu.edu/faculty/davise/papers/WinogradSchemas/WSChinese.html}} are available on the official webpage of the challenge.
Authors of French and Portugese translation both report having to make some changes to the content to avoid unintended cues, such as grammatical gender.
In the case of Portugese, $8$ sentences had to be dropped, as no appropriate translation could be found.

\subsection{Definite Pronoun Resolution Dataset}

The Definite Pronoun Resolution (\textsc{Dpr}) dataset is an easier variation of the Winograd Schema Challenge~\cite{DPR}.
The constraints on the Winograd schemas have been relaxed, and several examples in the dataset are not \textit{Google-proof}.
The dataset consists of $1322$ training examples and $564$ test examples, constructed manually.
$6$ examples in the training set reappear in \textsc{Wsc273} in a very similar form. 
These should be removed when training on \textsc{Dpr} and evaluating on \textsc{Wsc273}.
This dataset is also referred to as \textsc{WscR}, as named by Opitz and Frank~\shortcite{WSCRanking}.

An expanded version of this dataset, called \textsc{WinoCoref}, has been released by \citeauthor{WinoCoref}~\shortcite{WinoCoref}, who 
further annotate all previously ignored mentions (in their work, a mention can be either a pronoun or an entity) in the sentences that were not annotated in the original work.
In this way, they add $746$ mentions to the dataset, $709$ of which are pronouns.
Moreover, \citeauthor{WinoCoref}~\shortcite{WinoCoref} argue that accuracy is not an appropriate metric of performance on the \textsc{WinoCoref} dataset and introduce a new one, called \textit{AntePre}.

They define \textit{AntePre} as follows:
Suppose there are $k$ pronouns in the dataset, and each pronoun has $n_1, \ldots, n_k$ antecedents.
We can treat finding the correct candidate for each pronoun as a binary classification for each antecedent-pronoun pair.
Let $m$ be the number of correct binary decisions.
\textit{AntePre} is then computed as ${m}\,/\,{\scriptstyle{\sum_{i=1}^k n_i}}$.

\subsection{Pronoun Disambiguation Problem Dataset}

The Pronoun Disambiguation Problem (PDP) dataset consists of 122 problems of pronoun disambiguation collected from classic and popular literature,  newspapers, and magazines. Because constructing Winograd schemas according to Levesque’s original guidelines was a difficult, manual process, PDPs, which were collected and vetted rather than constructed, were intended to be used as a gateway set before administration of the Winograd Schema Challenge~\cite{planwsc}.  Each PDP, once collected,  was vetted (and sometimes modified) to ensure that like Winograd schemas, the problems were of the sort that humans use commonsense knowledge to disambiguate, and were ``Google-proof.’’  Although each PDP was vetted to remove examples where sentence structure would help find the answer, there was no ``special'' word, and thus, unlike Winograd schemas, no guarantee that sentence structure could not be exploited. PDPs were therefore expected to be easier than Winograd schemas.  

Example:
\textit{Do you suppose that Peter is responsible for the captain’s illness? Maybe he bribed the cook to put something in \textbf{his} food.}
\\
The referent of {\textbf{his}} 
is: (a) Peter or (b) the captain.

62 examples of PDPs were published before the Winograd Schema Challenge was administered~\footnote{\url{http://commonsensereasoning.org/disambiguation.html}}, and 60 PDPs were included in the Winograd Schema Challenge that was administered at IJCAI 2016~\footnote{\url{https://cs.nyu.edu/faculty/davise/papers/PDPChallenge.xml}}  \cite{WSC2016}.
A corpus of 400 sentences was collected semi-automatically from online text, with less vetting, by Davis and Pan~\shortcite{PDPlarge}\footnote{https://cs.nyu.edu/faculty/davise/annotate/corpus.xml}.


\subsection{Winograd Natural Language Inference Dataset}
\label{Wnli-section}
The Winograd Natural Language Inference (\textsc{Wnli}) dataset is part of the GLUE benchmark~\cite{GLUE} and is a textual entailment variation of the Winograd Schema Challenge.
An example from \textsc{Wnli} is given below with the goal to determine whether the hypothesis follows from the premise.

\smallskip 
\noindent\textbf{Premise: }\textit{The city councilmen refused the demon\-strators a permit because they feared violence.}

\noindent\textbf{Hypothesis: } \textit{The demonstrators feared violence.}

\noindent\textbf{Answer: } true / \textbf{false}

\smallskip 

The dataset consists of $634$ training examples, $70$ validation examples, and $145$ test examples.
Training and validation sets contain a major overlap with the \textsc{Wsc273} dataset, while test samples come from a previously unreleased collection of Winograd schemas.
Not all examples in this dataset contain the \textit{special} word and therefore do not come in pairs.
Kocijan \emph{et al.}~\shortcite{MaskedWiki} note that examples are much easier to approach if the Winograd schemas are transformed from the textual entailment back into the pronoun resolution problem, and approached as such.

The same collection of examples is used for the SuperGLUE benchmark~\cite{SuperGLUE} as a pronoun resolution problem to begin with.
For the purpose of this survey paper, \textsc{Wnli} and SuperGlue \textsc{Wsc} are considered the same dataset.
They consist of the same examples and all approaches to \textsc{Wnli} described in this paper transform the examples as noted in the previous paragraph.

\subsection{WinoGender Dataset}

Unlike the previous 
datasets, \textsc{WinoGender} was created as a diagnostic dataset and is aimed to measure gender bias of the systems for pronoun resolution~\cite{WinoGender}.
\textsc{WinoGender} consists of $120$ hand-written sentence templates, together with candidates and pronouns that can be inserted into the templates to create valid sentences.

In each sentence, one of the candidates is an occupation, usually one with a high imbalance in gender ratio (e.g., surgeon).
The other candidate is a participant (e.g., patient) or a neutral \textit{someone}.
For each sentence, either of the pronouns \textit{he, she}, or \textit{they} can be included to create a valid sentence, as the candidates are gender-neutral.
All together, this gives $720$ Winograd schemas.
An example from the dataset is 

\textit{The surgeon operated on the child with great care; [\textbf{his}/\textbf{her}] [tumor/affection] had grown over time.}

Note that the gender of the pronoun does not affect the expected answer; however, a biased system that associates the pronoun \textit{his} with the surgeon is likely to answer one of them incorrectly.
The aim of this dataset is not to measure model performance, as its data distribution is highly skewed, but to help analyse the models for gender bias.

\subsection{WinoBias Dataset}

\textsc{WinoBias} was created 
by Zhao \emph{et al.}~\shortcite{WinoBias}, which tries to identify gender bias in pronoun resolution models.
\textsc{WinoBias} and \textsc{WinoGender} were created concurrently but independently, despite the same objective.
They introduce a dataset with $3,160$ sentences, split equally into development and test.
Each sentence contains two candidates that are selected from a list of jobs with highly imbalanced gender ratio.

Two different templates are used to create Winograd schemas.
Type 1 sentences follow a structure that does not give away any syntactic cues. 
The authors thus estimate these sentences to be more challenging. 
An example of such a sentence is 

\textit{The farmer knows the editor because [\textbf{he}/\textbf{she}] [is really famous/likes the book].}

Type 2 sentences can be answered based on the structure of the sentence. 
The authors thus expect the models to perform better.
An example of such a sentence is 

\textit{The accountant met the janitor and wished [\textbf{her}/\textbf{him}] well.}

Its ``twin pair'' has the candidates swapped.
As the structure of the sentence gives the answer away, there is no \textit{special~word}.

Moreover, the authors evenly split the whole dataset into \textit{pro-stereotypical} and \textit{anti-stereotypical}, depending on whether the gender of the pronoun matches the most common gender of the referent occupation or not.
They observe that publicly available models for co-reference resolution exhibit a major difference (up to $21.1\%$ $F_1$) in performance on \textit{pro-} and \textit{anti-} subsets of the dataset.

\subsection{WinoGrande Dataset}

The \textsc{WinoGrande} dataset is a large-scale Winograd Sche\-ma Challenge dataset ($44k$ examples)~\cite{WinoGrande} collected via crowdsourcing on Amazon Mechanical Turk.
To prevent the crowd from creating lexically and stylistically repetitive examples, the workers are primed by a randomly chosen topic from a WikiHow article as a suggestive context.
Finally, the authors use an additional crowd of workers to ensure that the sentences are hard but not ambiguous to humans.
These measures were taken to ensure that there is no instance-level bias that models could exploit.

However, checking for instance-level cues is often not enough, as models tend to pick on dataset-level biases.
The authors additionally introduce the \textsc{AfLite} adversarial filtering algorithm.
They use a fine-tuned RoBERTa language model~\cite{Roberta} to gain contextualized embeddings for each instance.
Using these embeddings, they iteratively train an ensemble of linear classifiers, trained on random subsets of the data and discard top-$k$ instances that were correctly resolved by more than $75\%$ of the classifiers.
By iteratively applying this algorithm, the authors identify a subset ($12,282$ instances), called \textsc{WinoGrande}\textsubscript{debiased}.
Finally, they split this dataset into training ($9,248$), development ($1,267$), and test ($1,767$) sets.
They also released the unfiltered training set \textsc{WinoGrande}\textsubscript{all} with $40,938$ examples.

\subsection{WinoFlexi Dataset}
Similarly to WinoGrande, Isaak and Michael~\shortcite{WinoFlexi} aim to construct a dataset through crowdsourcing.
They build their own system and collect $135$ pairs of Winograd schemas ($270$ examples).
Unlike workers on \textsc{WinoGrande}, workers on \textsc{WinoFlexi} are not presented with any particular topic and are free to pick it on their own.
Despite this, authors find the collected schemas to have decent quality achieved through manual supervision between workers.
\section{Approaches to Winograd Schema Challenge}

At least three different methods have been used to try to solve the Winograd Schema Challenge. One class of approaches consists of feature-based approaches, typically extracting information such as semantic relations.
Additional \textit{commonsense knowledge} is usually included in form of explicitly written rules from knowledge bases, web searches, or word co-occurrences.
The collected information is then used to make a decision, using rule-based systems, various types of logics, or discrete optimization algorithms.
We observe that the extraction of relevant information from the sentence is usually the bottleneck of these approaches.
Given the nature of the challenge, even the slightest noise in the feature collection can make the problem unsolvable.

The second group of approaches are neural approaches, excluding language-model-based approaches, which we consider as a separate group.
Neural-network-based approaches usually read the sentence as a whole, removing the bottleneck of information extraction.
To incorporate background information, these networks or their components are usually pre-trained on unstructured data, usually unstructured text, or other datasets for coreference resolution.
Common approaches to the tasks in this group take advantage of semantic similarities between word embeddings or use recurrent neural networks to encode the local context.
We find this group of approaches to lack reasoning capabilities, as semantic similarity or local context usually do not contain sufficient information to solve Winograd schemas.

The third group includes approaches that make use of large-scale pre-trained language models, trained with deep neural networks, extensively pre-trained on large corpora of text.
Some of the approaches then additionally fine-tune the model on Winograd-Schema-Challenge-style data to maximize their performance.
Approaches in this group achieve visibly better performance than approaches from the first two groups.

Due to a scattered nature of the results, we decided not to combine them into one large table.
Not all methods are evaluated on the same set of examples.
Moreover, choices non-crucial to the idea, such as the choice of word embeddings or a language model can significantly affect the results, making the direct comparison unfair.

\subsection{Feature-based Approaches}

This section covers the approaches that collect knowledge in form of explicit rules from knowledge bases, internet search queries, and use logic-based systems or optimization techniques to deduce the answer.
We emphasize that results of methods that rely on search engines, such as Google, can be irreproducible, 
as they strongly depend on the search~results.

The first model was introduced by Rahman and Ng~\shortcite{DPR} together with the \textsc{Dpr} dataset.
The features that consist of Google queries, narrative chains, and semantic compatibility, were used to rank candidates with an SVM-based ranker.
Their approach achieved $73.05\%$ accuracy on the \textsc{Dpr} test set.
Peng \emph{et al.}~\shortcite{WinoCoref} achieved $76.41\%$ accuracy on this same dataset using integer linear programming and manually constructed schemas to learn conditioning from unstructured text.
They additionally evaluated their model on \textsc{WinoCoref}, where they achieved $89.32$ AntePre score.

Sharma \emph{et al.}~\shortcite{Kparser} constructed a general-purpose semantic parser and use it to parse and answer Winograd sche\-mas.
The parser is used to extract relevant information from the sentence and internet search queries.
Answer set programming (ASP)~\cite{ASP1,ASP2} is then used to define the rules and constructs for reasoning.
Due to their focus on specific types of reasoning, the authors only evaluate their approach on $71$ examples from \textsc{Wsc285} where such reasoning is present, with their systems correctly answering $49$ examples ($69\%$ accuracy).
As noted by Zhang and Song~\shortcite{DistributedWSC}, this same approach achieves $50\%$ accuracy on a different subset of $92$ examples.
Isaak and Michael~\shortcite{WinoFlexi} report this system to correctly solve $38\%$ of \textsc{WinoFlexi} examples, to incorrectly solve $36\%$, and to make no decision on the remaining examples.
As shown by Sharma~\shortcite{ASP3}, the sentence parsing and the knowledge collection are the bottleneck of this process.
Sharma~\shortcite{ASP3} develops an ASP-based algorithm, called \textsc{WiSCR}, which correctly solves $240$ out of $285$ \textsc{Wsc285} examples, if the input and background knowledge are provided by a human.
On the other hand, this same algorithm only solves $120$ of the examples, if it uses K-Parser for input parsing and a search engine for knowledge hunting.

Emami \emph{et al.}~\shortcite{KnowledgeHunter} developed the first model to achieve a better-than-chance accuracy ($57.1\%$) performance on the entire \textsc{Wsc273}.
Their system is completely rule-based and focuses on high-quality knowledge hunting, rather than reasoning, showing the importance of the former.
Unlike neural approaches from later sections, this model is not negatively affected by switching candidates.

Isaak and Michael~\shortcite{WinoSense} take a similar approach and use a collection of heuristics and external systems for text processing, information extraction, and reasoning.
The final system correctly resolves $170$ of the $286$ examples from an older collection of Winograd schemas\footnote{\url{https://cs.nyu.edu/faculty/davise/papers/OldSchemas.xml}} and $59\%$ of \textsc{WinoFlexi}.

An interesting approach to reasoning was proposed by F\"ahndrich \emph{et al.}~\shortcite{MarkerPassing}, who build a graph for each example by combining knowledge about words from several knowledge bases with semantic and syntactic information. 
They place a collection of markers on the pronoun and iteratively distribute them across the graph according to a manually designed set of rules.
The candidate with the greatest number of markers after $n$ steps is considered the answer.
The approach is evaluated on \textsc{Pdp}, where it obtains $74\%$ accuracy.

\subsection{Neural Approaches}
This section contains approaches that rely on neural networks and deep learning, but do not use pre-trained language models.
Models in this section are usually designed, built, and trained from scratch, while models that use language models are usually built on top of an off-the-shelf pre-trained neural network.
We find that several ideas introduced in this section are later adjusted and scaled to language models; see~Sec\-tion~\ref{section-LM}.
Note that each work comes with a collection of mo\-del-specific architecture designs that are not covered in~detail.

Liu \emph{et al.}~\shortcite{NAM} were the first to use neural networks to approach the challenge.
They introduce a neural association model to model causality and automatically construct a large collection (around $500,000$) of cause-effect pairs, that are used to train the model.
The model is then trained to predict whether the second part of the schema is the consequence of the first one.
For evaluation, Liu \emph{et al.}~\shortcite{NAM} manually select $70$ Winograd schemas from the \textsc{Wsc273} dataset that rely on cause-effect reasoning.
Their best model achieves $70\%$ accuracy on this selected subset.
In their subsequent work, Liu \emph{et al.}~\shortcite{KEE} extend this approach and use it at the Winograd Schema Challenge 2016~\cite{WSC2016}.
They develop their own pre-trained word embeddings, whose semantic similarity should correlate with cause-effect pairs, and train the final model on Ontonotes dataset for coreference resolution~\cite{Ontonotes}.
This method achieved the final score of $58.3\%$ on the \textsc{Pdp} dataset and $52.8\%$ on  \textsc{Wsc273}.

Zhang and Song~\shortcite{DistributedWSC} similarly try to augment word embeddings that can take advantage of dependencies in the sentence.
Unlike Liu \emph{et al.}~\shortcite{KEE}, their model is completely unsupervised and is not additionally trained on any labelled data.
They modify the Skip-Gram objective for word embedding pre-training to additionally use and predict semantic dependencies, which can thus be used as additional information at test time.
The introduced approach is tested on a manually selected set of $92$ \textit{easy} Winograd schemas from the \textsc{Wsc273} dataset, achieving a $60.33\%$ accuracy.
Wang \emph{et al.}~\shortcite{UDSSM} take a step further with the unsupervised deep semantic similarity model (UDSSM).
Instead of augmenting the word embedding, they train BiLSTM modules to compute contextualized word embeddings.
The best performing ensemble of their models achieves $78.3\%$ accuracy on \textsc{Pdp} and $62.4\%$ accuracy on \textsc{Wsc273}.

Opitz and Frank~\shortcite{WSCRanking} are the first to try to generalize from \textsc{Dpr} to \textsc{Wsc273} by training on the former and testing on the latter.
We note that authors do not mention removing the overlap between them. 
In their approach, they replace the pronoun in question with one of the candidates.
They design several Bi-LSTM-based models and train them to rank the sentence with the correct candidate better than the sentence with the incorrect candidate. 
Their best approaches achieve $63\%$ on \textsc{Dpr} and an accuracy of $56\%$ on \textsc{Wsc273}, showing that generalizing from \textsc{Dpr} to \textsc{Wsc273} is not trivial.

\subsection{Language Model Approaches}
\label{section-LM}
This section covers the approaches that use neural language models to tackle the Winograd Schema Challenge.
Most of them use one or more language models that were trained on a large corpus of text.
Several authors use large pre-trained language models, such as BERT~\cite{Bert}, and have to tailor their approach accordingly.
Many works thus focus on better fine-tuning of such language models instead of inventing new architectures.

Trinh and Le~\shortcite{WinogradGoogle} were the first to use pre-trained language models.
Similarly to Opitz and Frank~\shortcite{WSCRanking}, two sentences are created from each example by replacing a pronoun with each of the two candidates.
A language model, implemented as an LSTM and pre-trained on a large corpus of text is used to assign them a probability.
The ensemble of $14$ such language models obtained was evaluated on the \textsc{Pdp} ($70\%$ accuracy) and \textsc{Wsc273} datasets ($63.74\%$ accuracy).
Trichelair \emph{et al.}~\shortcite{WSCAnalysis} have shown that this ensemble is highly inconsistent in the case of swapped candidates and mainly works well on the associative subset of \textsc{Wsc273}.
Radford \emph{et al.}~\shortcite{GPT2} apply the same method to evaluate their GPT-2 language model and achieve $70.7\%$ accuracy on \textsc{Wsc273}.
Melo \emph{et al.}~\shortcite{WSCportugese} use this method on their Portugese version of the Winograd Schema Challenge.
They use an LSTM-based language model, trained on text from Portugese Wikipedia, but only achieve a chance-level performance.

Prakash \emph{et al.}~\shortcite{KnowledgeHuntingLMs} extend this approach with knowledge hunting.
They find sentences on the web that describe a similar situation, but may be easier to resolve.
They assume that the pronoun refers to the same candidate.
They use the same method as Trinh and Le~\shortcite{WinogradGoogle} to compute probabilities of each candidate for each pronoun.
The assumption that the pronoun in the mined sentence and the pronoun in the Winograd schema refer to the same entity is described and imposed with probabilistic soft logic~\cite{PSL}.
That is, all pronouns are resolved to the same candidate in the most probable way.
The best model obtained by combining language models and knowledge hunting in this way achieves $71.06\%$ accuracy on \textsc{Wsc273} and $70.17\%$ accuracy on \textsc{Wsc285}.

Klein and Nabi~\shortcite{AttentionWSC} analyse inner attention layers of a pre-trained BERT-base language model~\cite{Bert} to find the best referent.
They define a \textit{maximum attention score}, which computes how much the model has attended to each candidate across all the layers and attention heads.
The model is evaluated on both \textsc{Pdp} ($68.3\%$ accuracy) and \textsc{Wsc273} ($60.3\%$ accuracy).

Kocijan \emph{et al.}~\shortcite{MaskedWiki} adapt scores from Trinh and Le~\shortcite{WinogradGoogle} to masked language models, i.e., BERT~\cite{Bert}.
They additionally introduce an unsupervised pre-training dataset \textsc{MaskedWiki} from English Wikipedia, which is constructed by masking repeated occurrences of nouns ($130$M examples, downscaled to $2.4$M).
When fine-tuned on both \textsc{MaskedWiki} and \textsc{Dpr}, BERT-large achieves $72.5\%$ performance on \textsc{Wsc273} and $74.7\%$ on \textsc{Wnli}.
By transforming \textsc{Wnli} examples as introduced in Section~\ref{Wnli-section}, this was the first model to beat the majority-class baseline.

Several authors have used this approach to \textsc{Wnli} as part of the GLUE benchmark~\cite{GLUE} with best performance achieved by T5 at $94.5\%$~\cite{T5}.
The improvement over Kocijan \emph{et al.}~\shortcite{MaskedWiki} usually comes from more extensive pre-training, and training on the training set of \textsc{Wnli}, which was not used by Kocijan \emph{et al.}~\shortcite{MaskedWiki}, due to its overlap with \textsc{Wsc273}.

In their subsequent work, Kocijan \emph{et al.}~\shortcite{WikiCREM} introduce a dataset called \textsc{WikiCREM} ($2.4$M examples), generated in the same way as \textsc{MaskedWiki}, but restricted to masking personal names.
By pre-training on \textsc{WikiCREM} and fine-tuning on other coreference datasets they achieve $84.8\%$ accuracy on \textsc{Dpr}, $71.8\%$ on \textsc{Wsc273}, $74.7\%$ on \textsc{Wnli}, and $86.7\%$ on \textsc{Pdp}.

Ye \emph{et al.}~\shortcite{AMS} introduce an align, mask, and select (AMS) pre-training method for masked language models.
They find sentences that contain entities that are directly connected in the ConceptNet knowledge base~\cite{ConceptNet}.
They mask one of them and train the model to pick it from a list of candidates over other similar candidates.
They fine-tune the obtained model in the same way as Kocijan \emph{et al.}~\shortcite{MaskedWiki} to achieve $75.5\%$ and $83.6\%$ accuracy on \textsc{Wsc273} and \textsc{Wnli}, respectively.

He \emph{et al.}~\shortcite{HNN} combine the masked token prediction model by Kocijan \emph{et al.}~\shortcite{MaskedWiki} with the semantic similarity model by Wang \emph{et al.}~\shortcite{UDSSM} to create a hybrid neural network model.
The combined model achieves $75.1\%$ accuracy on \textsc{Wsc285}, $90.0\%$ accuracy on \textsc{Pdp}, and $89.0\%$ on \textsc{Wnli}.
The \textsc{Wnli} result was achieved by using an ensemble.

A different use of the BERT language model was used by Ruan \emph{et al.}~\shortcite{WSCNSP}, who take advantage of the BERT \textit{next sentence prediction} feature.
In addition to replacing the pronoun with a candidate, they split the sentence into two parts, predicting whether the second part semantically follows the first one.
To improve the performance, Ruan \emph{et al.}~\shortcite{WSCNSP} encode syntactic dependency by changing some of the attention tensors within BERT and train on \textsc{Dpr}.
The BERT-large model combined with all the features achieves $71.1\%$ accuracy on the  \textsc{Wsc273} dataset.



Sakaguchi \emph{et al.}~\shortcite{WinoGrande} use the same approach when evaluating the RoBERTa baseline for the \textsc{WinoGrande} dataset; however, they do not modify any attention layers, and train on \textsc{WinoGrande} rather than \textsc{Dpr}. 
They report achieving $79.1\%$ accuracy on \textsc{WinoGrande}, $90.1\%$ accuracy on \textsc{Wsc273}, $87.5\%$ on \textsc{Pdp}, $85.6\%$ on \textsc{Wnli}, and $93.1\%$ on \textsc{Dpr}.
To this point, this is the highest performance achieved on the \textsc{Wsc273} dataset by a large margin, showing the impact of additional training data.
Curiously, they report achieving chance-level improvement on the validation set of \textsc{WinoGrande} when training on the \textsc{WinoGrande}\textsubscript{debiased}.
They suspect that the introduced model performed well on \textsc{WinoGrande}\textsubscript{full}, because it trained to exploit a systemic bias within the dataset.

\section{Conclusion}

In this paper, we have reviewed and compared the datasets of Winograd
schemas that have been created and the many systems that have been developed
to attempt to solve them.  Currently, the best of these systems, which exploit deep neural networks and incorporate very large and sophisticated pre-trained transformer models, such as BERT or RoBERTa finetuned, are able to
achieve around 90\% accuracy rates on \textsc{Wsc273} and similar datasets.

Levesque \emph{et al.}~\shortcite{WinogradSchema} claimed that 
because of the use of twin sentences,
``clever tricks involving
word order or other features of words or groups of words {\em will not work}
[emphasis added].'' This prediction has been falsified, at least as far as
the dataset produced with that paper is concerned. The paper did not anticipate the power of neural networks,
the rapid advances in natural language modelling
technology resulting in language models like BERT, and the subtlety and complexity of the patterns in words that such technologies would be able to find and apply.

The systems that have succeeded on the Winograd
Schema Challenge have succeeded on the pronoun disambiguation task in small passages of text, but they have not demonstrated either the ability to perform other natural language understanding tasks, or common sense. They have not demonstrated the ability to reliably answer simple questions about narrative text~\cite{RAI} or to answer simple questions about everyday situations. Similarly, text generated using even state-of-the-art language modeling systems, such as GPT-2, frequently contains incoherences~\cite{NextDecadeInAI}. 


The commonsense reasoning and the natural language understanding communities require new tests, more probing than the Winograd Schema Challenge, but still easy to administer and evaluate. Several tests have been proposed and seem promising. The problem of tracking the progress of a world
model in narrative text is discussed by Marcus~\shortcite{NextDecadeInAI}.
A number of proposed replacements
for the Turing Test~\cite{BeyondTT} likewise draw heavily
on various forms of commonsense knowledge. 
\section*{Acknowledgements}
This work was supported by the EPSRC studentship    OUCS/EPSRC-NPIF/VK/1123106.

\end{document}